\documentclass[runningheads]{llncs}
\usepackage{graphicx}
\usepackage{subfig}
\usepackage[export]{adjustbox}
\usepackage{siunitx}
\usepackage{dirtytalk}
\usepackage{amsmath, amsfonts}
\usepackage{xcolor}
\usepackage[hyphens]{url}
\usepackage{multirow,booktabs}
\usepackage{hyperref}
\newcommand{\figref}[1]{Figure~\ref{#1}}
\newcommand{\tbref}[1]{Table~\ref{#1}}
\newcommand{\secref}[1]{Section~\ref{#1}}

\begin{document}
%
\title{SAFFIRE: System for Autonomous Feature Filtering and Intelligent ROI Estimation}
\titlerunning{SAFFIRE}
%
\author{Marco Boschi\inst{1}\thanks{Corresponding author.} \and
Luigi Di Stefano\inst{1} \and
Martino Alessandrini\inst{2}}
\authorrunning{M. Boschi et al.}
%
\institute{
Department of Computer Science and Engineering, University of Bologna\\
Viale Risorgimento 2, 40136 Bologna, Italy\\
\email{marco.boschi5@studio.unibo.it}, \email{luigi.distefano@unibo.it}
\and
Datalogic S.r.l.\\
Via San Vitalino 13, 40012 Calderara di Reno BO, Italy\\ \email{martino.alessandrini@datalogic.com}}
%
\maketitle              
\begin{abstract}
This work introduces a new framework, named SAFFIRE, to automatically extract a dominant recurrent image pattern from a set of image samples. Such a pattern shall be used to eliminate pose variations between samples, which is a common requirement in many computer vision and machine learning tasks. 

The framework is specialized here in the context of a machine vision system for automated product inspection. Here, it is customary to ask the user for the identification of an anchor pattern, to be used by the automated system to normalize data before further processing. Yet, this is a very sensitive operation which is intrinsically subjective and requires high expertise. Hereto, SAFFIRE provides a unique and disruptive framework for unsupervised identification of an optimal anchor pattern in a way which is fully transparent to the user. 

SAFFIRE is thoroughly validated on several realistic case studies for a machine vision inspection pipeline.

\end{abstract}
\section{Introduction}
Machine vision systems are key for the automation of the production line. Hereto, in a typical installation, a camera (or a set of cameras) is mounted on top of the conveyor transporting the manufactured items. The images are analyzed by a specific software which grades the \say{quality} of the product. Typical examples include: reading/verifying a printed text/barcode; identifying a pattern; measuring a part; verifying the absence of defects (flaw detection), and combinations of thereof. The output grading is then used to keep or discard that item for distribution.


It is often the case that the content to be analyzed resides in a small sub-portion of the image (i.e., a \emph{region of interest} (ROI)). As such, a necessary pre-processing step is the identification of such a region in the image. We call this process \say{ROI finding}. Of note, ROI finding is not only required for speedup, but is often necessary for the meaningfulness of the following analysis. For instance, one might need to select the correct line of text before reading it, or to locate the specific mechanical part to be measured. In many cases, the inspected items are left free to assume an arbitrary pose relative to camera and, as such, the ROI has to be tracked by the automated software independently on each sample.

A typical way to address the ROI finding problem is by instructing the system with an \emph{anchor pattern}. An anchor is a distinctive recurrent (i.e., present on all samples) image pattern (e.g., a specific mark on a printed label) identifying a local coordinate system where the ROI position is (as much as possible) constant. Such a pattern shall be detected by the machine vision system and used to position the ROI on a new sample. Importantly, the anchor is generally not located inside the ROI. Indeed, the focus of an inspection is commonly on regions which are expected to change. 

In many machine vision software products, the anchor pattern is annotated manually by the user when training the system. Manual annotation is disadvantageous for several reasons:
\begin{enumerate}
    \item It requires expertise: The user has to know what an anchor is and how to use it to configure the specific inspection (e.g., by instructing a template/pattern matching routine to locate the pattern on new samples);
    \item Visual identification of a viable anchor pattern can be complex. For instance, the anchor pattern has to be isolated among many (e.g., for complex printed labels) or, possibly, the anchor is represented by a set of visual features which are distributed sparsely in the image; 
    \item It is time consuming: It requires accurate visual inspection of multiple items in the lot;
    \item It is subjective: Different users might identify different anchors based on their expertise;
    \item The anchor selected by a human is very unlikely to be the \say{best} anchor usable by the ROI selection algorithm.
\end{enumerate}
Conditions (1), (2) and (3) are especially undesirable in a market which is constantly struggling to shift towards plug-and-play solutions.

This paper introduces a new framework, named SAFFIRE,\footnote{Pronounced as \emph{sapphire}.} which overcomes the aforementioned limitations:
\begin{itemize}
    \item The anchor is learned automatically from a set of training images;
    \item The framework requires only straightforward supervision which only implies a basic understanding of the task of interest. Namely, the user has to draw a ROI around the content of interest. Of note, this is considerably easier than identifying an anchor which can be uncorrelated with the specific task and distributed anywhere in the image;
    \item The computed anchor is the best pattern for the ROI selection algorithm.
\end{itemize}

The proposed framework presents also several computational advantages against alternatives based on Deep Neural Networks (DNN), which are rapidly gaining popularity:
\begin{itemize}
    \item SAFFIRE can be trained on a normal device within seconds (at most) while DNN-based systems would take minutes to hours and dedicated expensive hardware; 
    \item SAFFIRE requires very few train samples (order of units) while DNN-based systems would require thousands;
    \item SAFFIRE is agnostic to the specific task while DNN-based solutions are task-specific (e.g., text detectors \cite{text_detection_review});
    \item SAFFIRE can be trained incrementally, in that a new sample can be added seamlessly to the train set to refine the anchor pattern, while DNN-based solutions would require re-training the entire model (almost) from scratch;
    \item SAFFIRE outputs a lightweight anchor pattern which can be found on a new sample within tens of milliseconds. On the contrary, DNN-based systems output heavy models and inference operates at a slow rate;
\end{itemize}

The paper proceeds as follows: Related work is presented in \secref{sec:related_work}; \secref{sec:material_and_methods} describes the SAFFIRE algorithm, both at train-time and at run-time;
\secref{sec:evaluation_setup} presents the evaluation setup while \secref{sec:results} presents the results. \secref{sec:discussion} elaborates on the results and \secref{sec:conclusions} concludes the work.

\section{Related Work}
\label{sec:related_work}
At its core, SAFFIRE exploits an \emph{object detection} paradigm. Object detection is a well known problem in computer vision. Currently, the established approach is based on matching descriptors, i.e. compact representations of local image features, such as corner points, segments or blobs, extracted from the object of interest (see, e.g., \cite{sift_article}). Hereto, a range of techniques, such as the Generalized Hough Transform (GHT) or Random Sample Consensus (RANSAC) can be employed to retrieve instances of a \emph{query} pattern in a new image \cite{sift_article}.

Yet, applying the object detection framework as presented above requires the query pattern to be known. 
DNN-based approaches like YOLO \cite{yolo_v1} or Faster R-CNN \cite{faster_rcnn} may be used to implicitly learn such pattern, but they do not represent a solution that fulfills the constraints of the addressed application: only very few training images are available, the training time should be quite short and the processing speed at run-time must be as high as possible with limited computational resources.
In this context, we propose an original technique where the query pattern (i.e., the anchor) is learned autonomously from a set of images, satisfying the presented constraints without the employment of DNN-based solutions.
Technically, this is done by finding the shortest path in a directed acyclic graph (DAG) where each path identifies a possible model candidate.

Moreover, different descriptors are known to perform differently on different kinds of images. For instance, SIFT is arguably the most popular approach \cite{sift_article} for textured objects, although a number of alternative solutions have been also proposed, such as ORB \cite{orb_article} and SURF \cite{surf_article}. However, approaches based on segments \cite{lsd_article} and combination of segments \cite{bold_article} have been shown to be more appropriate for texture-less objects.

The choice of the best descriptor for the specific dataset is typically the result of a trial-and-error process. On the contrary, SAFFIRE implements a selection logic where the features most suited to the specific data are computed at train time.

Finally, classic feature matching is used at run-time to retrieve the anchor model in a new sample and re-position the ROI. In the case of multiple instances, SAFFIRE selects the solution where the image content inside the ROI is the closest to that learned on the train set.

\section{Material and Methods}
\label{sec:material_and_methods}

\begin{figure}[t]
	\centering
	\subfloat[Input: a set of labeled images.\label{fig:flowchart_train_in}]{
		\includegraphics[page=1,scale=1.2]{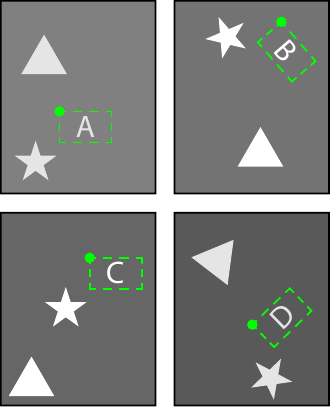}
	}\quad
	\subfloat[Feature extraction and matching.\label{fig:flowchart_train_1}]{
		\includegraphics[page=2,scale=1.2]{figures/algorithm.pdf}
	}\quad
	\subfloat[Testing model candidates and graph construction.\label{fig:flowchart_train_2}]{
		\includegraphics[page=3,scale=1.4]{figures/algorithm.pdf}
	}\quad
	\subfloat[Path finding.\label{fig:flowchart_train_3}]{
		\includegraphics[page=4,scale=1.4]{figures/algorithm.pdf}
	}\quad
	\subfloat[Output: model and model ROI.\label{fig:flowchart_train_out}]{
		\includegraphics[page=5,scale=1.5]{figures/algorithm.pdf}
	}
	\caption{Train mode: Build model and model ROI from train data. Please refer to the main text for a detailed explanation.}
	\label{fig:flowchart_train}
\end{figure}

The processing workflow of SAFFIRE at train time and test time is represented in \figref{fig:flowchart_train} and \ref{fig:flowchart_test}, respectively. A description of the pipeline follows.

\subsection{Training Data and Reference Image}
\label{sec:train_data}
A set of $N_{\text{train}}$ annotated train images $\{I_{0}, \ldots, I_{N_{\text{train}}-1}\}$ is provided as input (\figref{fig:flowchart_train_in}). One of such images, by convenience $I_{0}$, is adopted as \emph{reference} image. Each image is annotated with an oriented rotated rectangular ROI $R_i$. Note that orientation is in general relevant for the task (for instance, it can disambiguate text reading direction).

We assume that all train images contain a distinctive \emph{anchor} pattern mixed with a set of \emph{noise} patterns which are irrelevant for the identification of the ROI. The ultimate goal of the training phase is therefore to retain the anchor pattern while filtering irrelevant patterns out.

In the toy example in \figref{fig:flowchart_train_in}, the dashed rectangle denotes the ROI enclosing the \say{text} to be read. The highlighted corner is the origin of the ROI and indicates its orientation in that corners are ordered clockwise starting from the origin. The star represents the anchor pattern since its position relative to the ROI is stable. The triangle represents any other pattern in the image which cannot work as an anchor for this task.

\subsection{Feature Extraction}
\label{sec:feature_extraction}
A set of visual \emph{features} is extracted from all train images (the red and yellow dots in \figref{fig:flowchart_train_1}). Specifically, we use a well consolidated paradigm for object detection based on matching sets of visual landmarks using their local image descriptors \cite{feature_survey}.
We make use of multiple types of features to account for textured and texture-less patterns and the logic to choose the best feature type for the specific data is presented in \secref{sec:best_feature_type}. 

With this in mind, we define the anchor pattern as the largest subset of features such that:
\begin{itemize}
    \item It is present in all of the train images;
    \item Its position in different train images can be related by a geometrical transformation. Hereto, although general transformations can be used, we made use of similarities (i.e., roto-translations with isotropic scaling) since they were appropriate for the geometry of the experimental setup and relevant to industrial applications;
    \item Its position correlates best with the position of the ROI. More quantitatively, the overlap between the ROI's is maximal when they are expressed in the anchor's coordinate system (see \secref{sec:graph_construction}).
\end{itemize}
The procedure used to derive such an anchor pattern is described in the remaining of this Section.

The $j$-th feature on the $i$-th train image is denoted by $f_{i}^j$. A feature is geometrically defined by a position $\mathbf{x}_i^j$,
a direction $\theta_i^j$ and a size $l_i^j$. The image content around a feature is accounted for by a \emph{descriptor} $\mathbf{d}_i^j$.
Descriptors shall be used to match features between images, as denoted by the blue connecting lines in \figref{fig:flowchart_train_1}.

\subsection{The Training Graph}
\label{sec:graph_construction}
For each feature $f_{0}^j$ in the reference image, we find its 3 closest matches in the $i$-th train image. We use 3 nearest neighbours to allow good robustness to multiple instances and/or periodic patterns. We denote by $f_i^{k}$ one of such matches. Matches with a large inter-descriptor distance are discarded by applying Lowe's ratio test \cite{sift_article}. Further filtering is applied by using the following geometrical considerations:
\begin{itemize}
    \item Since in our setup scale is not expected to vary significantly, we exclude matches whose scale factor $s=l_0^j/l_i^k$ is not within the interval $[0.85,1.15]$;
    \item Recalling our definition of an anchor pattern (see \secref{sec:feature_extraction}), we exclude matches which are not useful to superimpose the ROI's between samples. Hereto, we use the (unique) spatial transform between $f_0^j$ and $f_i^k$ to bring $R_0$ and $R_i$ to a common origin, and we discard the match if the \emph{intersection over union} (IoU) between the transformed ROI's is below 0.2.
\end{itemize}

Matches voting for the same spatial transform $T_i^t$ are then aggregated into $N_i$ clusters, with $t=0,\ldots,N_i-1$.
Clustering is done by iterative application of the RANSAC algorithm \cite{ransac_article}. At each iteration, RANSAC is applied to find a transform $T_i^t$ and the \say{inliers} are removed from the list of matches. The procedure is repeated incrementally until not enough matches are left to estimate a transform.

As such, a directional acyclic graph (DAG), which we refer to as the \emph{training graph} (see \figref{fig:flowchart_train_2}), is created with a number of layers equal to the number of training images. Each layer contains $N_i$ nodes, one for each cluster computed as above. Each node $n_i^t$ in the $i$-th layer contains:
\begin{itemize}
    \item A candidate transform $T_i^t$ between $I_0$ and $I_i$;
    \item The indexes of the features in $I_0$ and $I_i$ voting for $T_i^t$.
\end{itemize}

The root layer in the training graph is obtained by matching the reference image with itself. By construction, this gives a single node described by an identity matrix and the indexes of all features in $I_0$.

\subsection{Search of Optimal Path in the Training Graph}
\label{sec:path_search}
As mentioned, the goal is to find the largest subset of features in $I_0$ which can be used as a common coordinate system maximizing the overlapping between the training ROI's.
The problem is solved by searching for the shortest root-to-leaf path in the training graph (in red in \figref{fig:flowchart_train_3}).
The length (cost) of a path is defined as a function of two terms: the former depends on the number of overlaps along the path, which is computed from the intersection of the features voting for all the transforms in the path; the latter depends on the alignment and overlap between the aligned ROI's after bringing them all to a common coordinate system by means of the transformations along the path.

To measure ROI alignment, we use the \emph{oriented IoU} ($o\mathtt{IoU}$), a modified standard \emph{intersection over union} (IoU) accounting for ROI orientation.
Indeed, the classical IoU between two rectangular shapes does not distinguish $\ang{180}$ flips around the center.
Specifically, for a node on the $n$-th layer, it is computed by using the set of spatial transformations on the intermediate path to first bring the $n$ ROI's to a common coordinate system.
The proposed oriented IoU is defined in such a way that its value for two flipped and perfectly overlapped rectangles is $0$, while their standard IoU would be $1$.

Having defined the cost of a path, we then use a tree-search algorithm to retrieve the optimal anchor pattern, the one corresponding to the path of least cost.

\subsection{The Anchor}
\label{sec:trained_model}
The best path is used to compute the final anchor model, which will be used at run-time to position the ROI on a new sample (see \figref{fig:flowchart_train_out}). Such model is composed of four parts: i) a set of \emph{model features}; ii) the associated \emph{model descriptors}, iii) a \emph{ROI model} $R_m$ and iv) a \emph{ROI descriptor}.

\emph{Model features} are the subset of $\{f_0\}$ which vote for all the transforms in the best path. For what concerns the \emph{model descriptors}, although one could use the descriptors computed on the reference image alone, this would result in a high bias to the specific choice of the reference image.
As such, we defined descriptors which also account for the statistical variation of image content between train samples.

The \emph{ROI descriptor} is used to keep track of the image content inside the ROI. It will be used to discriminate between multiple instances of the anchor model at run time, as described in \secref{sec:testing}. Let's note that a \emph{global} image descriptor, i.e. accounting for average image properties such as mean image value or average contrast, is best suited for machine vision applications. Indeed, fine image details can change between samples (for instance, the content of a printed text).
As for the model descriptors, the ROI descriptor is not tailored to a specific train image, but is rather based on the statistical distribution of image content between images.

\subsection{Selection of the Best Feature Type}
\label{sec:best_feature_type}
Since different features are best suited for different patterns (e.g., textured vs. texture-less objects), SAFFIRE employs an optimized feature selection strategy which chooses the feature-descriptor pair which performs best on the specific data. Both accuracy and speed are used as a selection criterion.

\subsection{Testing}
\label{sec:testing}
\begin{figure}
	\centering
	\subfloat[Input: a test image.\label{fig:flowchart_test_in}]{
		\includegraphics[page=6,scale=1.5]{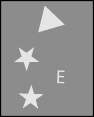}
	}\quad
	\subfloat[Search model in image.\label{fig:flowchart_test_1}]{
		\includegraphics[page=7,scale=1.5]{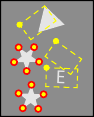}
	}\quad
	\subfloat[Remove bad matches with ROI content.\label{fig:flowchart_test_2}]{
		\includegraphics[page=8,scale=1.5]{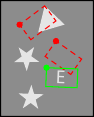}
	}\quad
	\subfloat[Output: ROI on the new image.\label{fig:flowchart_test_out}]{
		\includegraphics[page=9,scale=1.5]{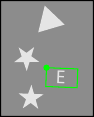}
	}
	\caption{Test mode: Find the trained model on unseen data. Please refer to the main text for a detailed explanation.}
	\label{fig:flowchart_test}
\end{figure}

The trained model is used at run-time to position the ROI in a new unseen sample (see \figref{fig:flowchart_test_in}) with a two-step process.

First, a standard object detection paradigm is used to find instances of the trained model in the new image (\figref{fig:flowchart_test_1}). Namely, features and descriptors of the selected type (see \secref{sec:best_feature_type}) are extracted on the test image and matched to the anchor pattern model (see \secref{sec:graph_construction}). Using the GHT, as in \cite{sift_article}, feature correspondences are turned into votes to an accumulation array (AA) with peaks in the AA representing possible instances of the anchor model in the test image. Hereto, a peak pruning procedure, based on the relative number of votes in each peak, is applied to keep only the most significant matches. For each valid instance of the anchor model, the corresponding spatial transform is used to position the ROI model in the new sample, as described in see \secref{sec:trained_model}.

With regard to \figref{fig:flowchart_test}, the test image contains two instances of the \say{star} anchor model, which presents circular symmetry. This could result in many matches (hence, ROI positions) with equal score, but only that containing the letter \say{E} is relevant. Such matches are displayed in \figref{fig:flowchart_test_1} as yellow dashed boxes.

The following step consists therefore in discarding non-relevant matches (see \figref{fig:flowchart_test_2}) by computing the content descriptor for each candidate ROI (see \secref{sec:trained_model}). The weighted $\ell$-2 distance to the model ROI descriptor $\mathbf{d}_{\mathrm{model}}$ is:
\begin{equation}
d(\mathbf{d}_{\mathrm{test}}, \mathbf{d}_{\mathrm{model}}, \mathbf{w}) = \sqrt{\sum_{i=0}^{47} w_i \left( \mathbf{d}_{\mathrm{test}}(i) - \mathbf{d}_{\mathrm{model}}(i) \right)^2}
\end{equation}
where $\mathbf{d}(i)$ denotes the $i$-th bin of $\mathbf{d}$. A weight $w_i \in [0,1]$ assigns highest (respectively, lowest) weight to bins with the lowest (respectively, highest) variance. The ROI with the lowest distance is the one used to analyze the sample at run time, as depicted in \figref{fig:flowchart_test_2} and \figref{fig:flowchart_test_out}.  

\section{Evaluation Setup}
\label{sec:evaluation_setup}

\subsection{Evaluation Datasets}

\begin{figure}
    \centering
    \subfloat[StarCart, toy dataset.\label{fig:starcart_ds}]{\includegraphics[width=.23\textwidth,cframe=gray!40]{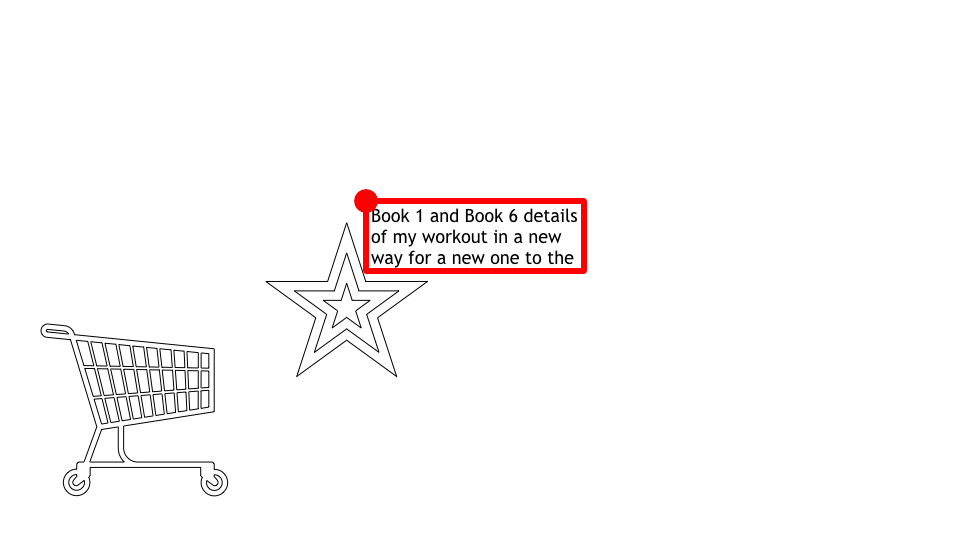}\includegraphics[width=.23\textwidth,cframe=gray!40]{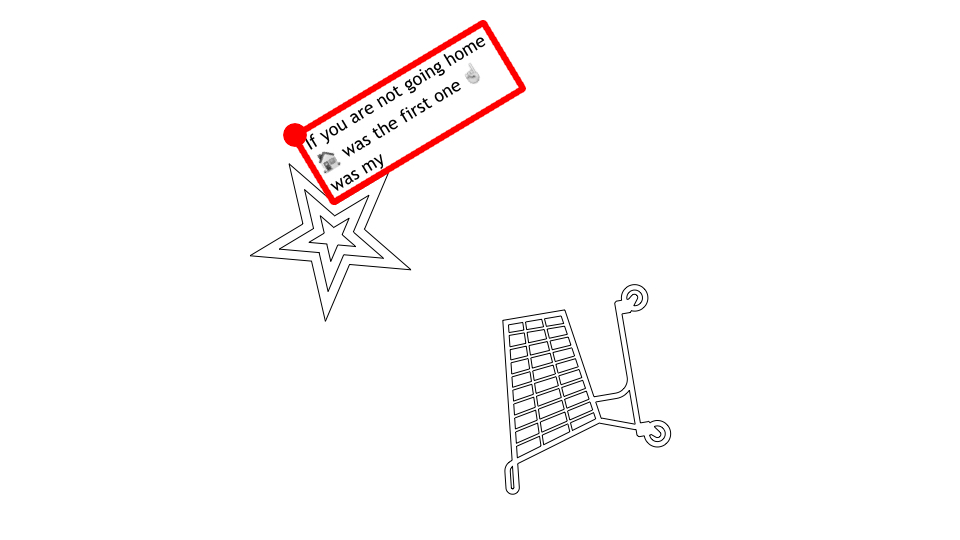}}\quad
    \subfloat[Cans, 87 images, larger size.]{\includegraphics[width=.23\textwidth,cframe=gray!40]{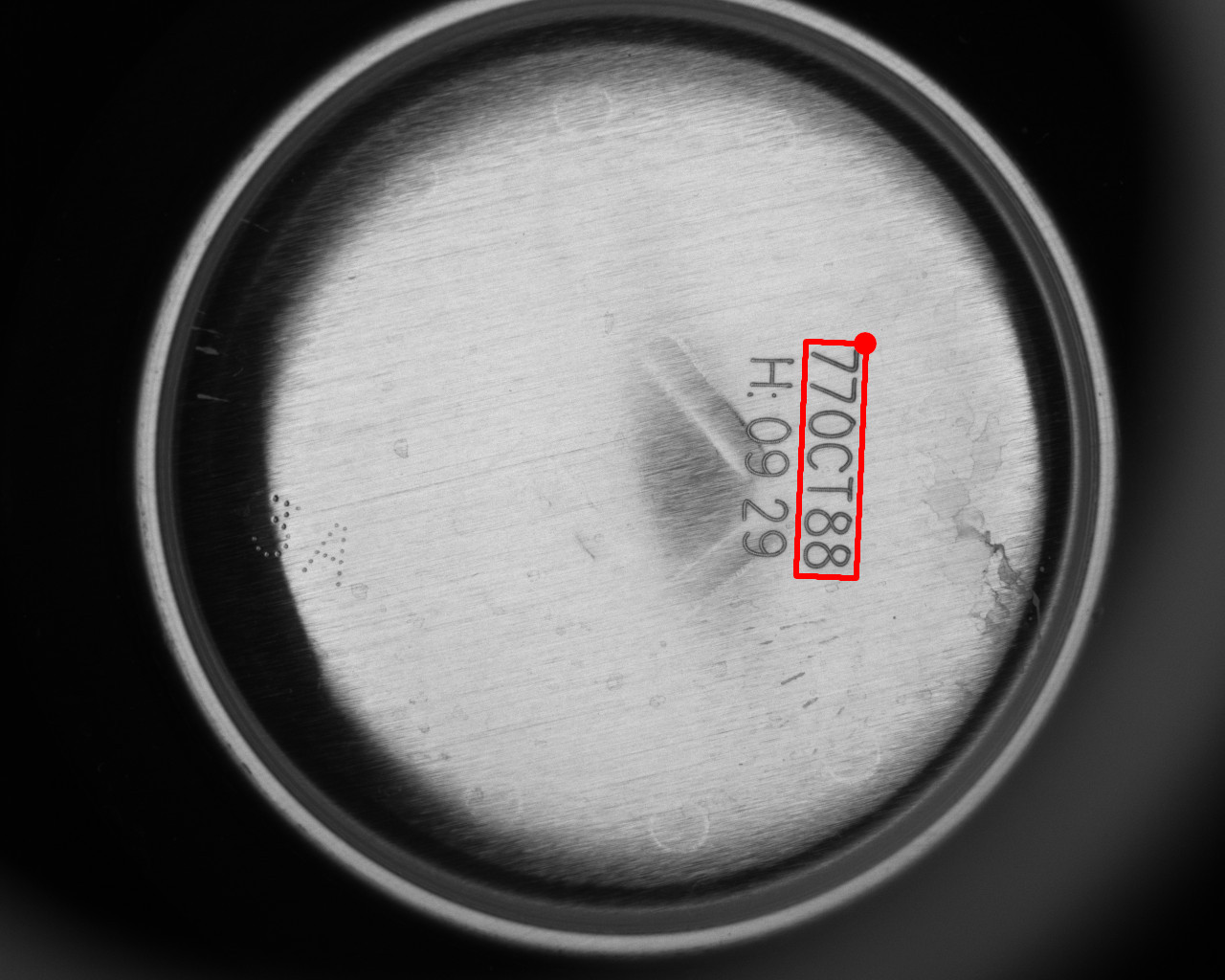}\includegraphics[width=.23\textwidth,cframe=gray!40]{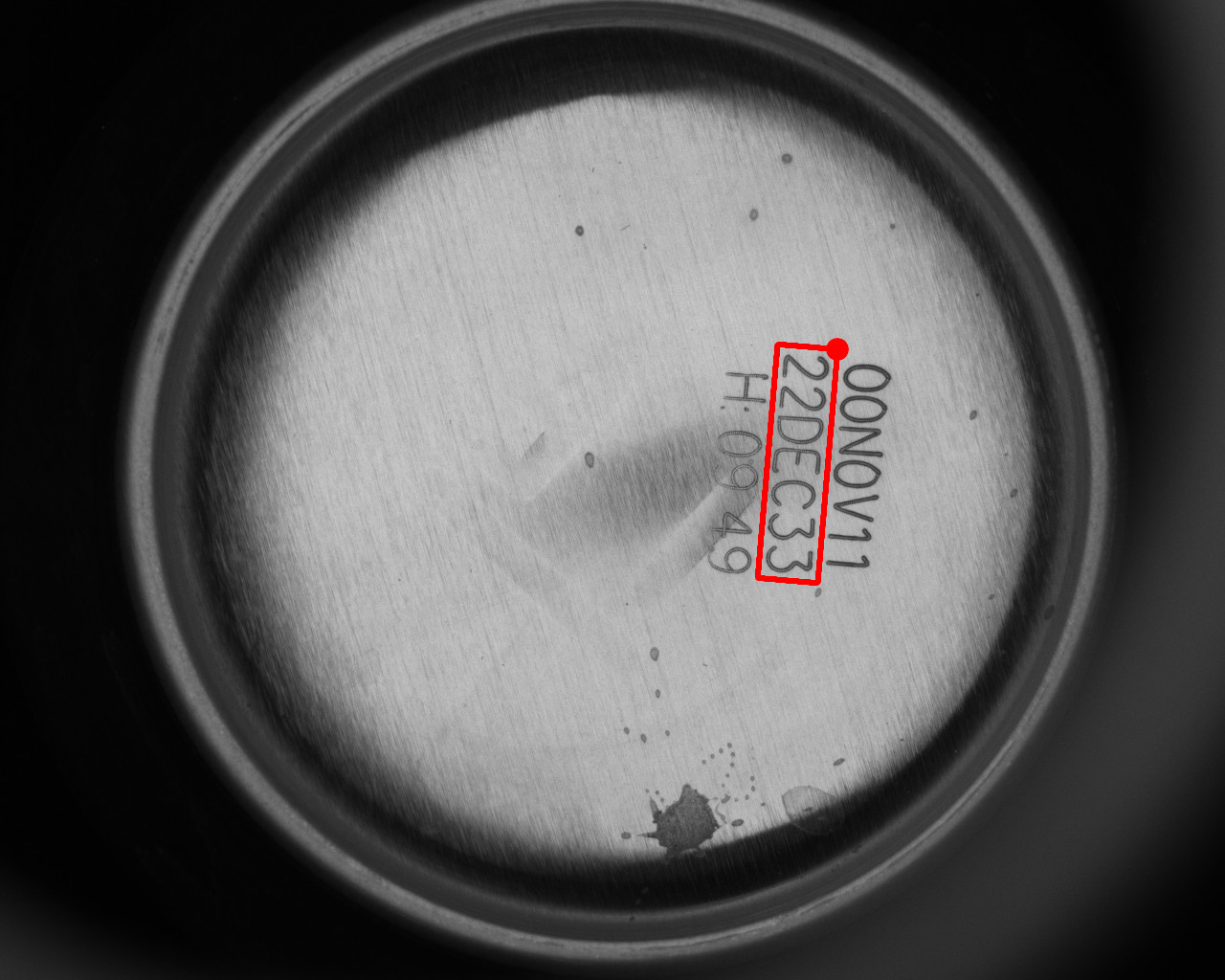}}\\
    \subfloat[Curry, 16 images, larger size.\label{fig:curry_ds}]{\includegraphics[width=.23\textwidth,cframe=gray!40]{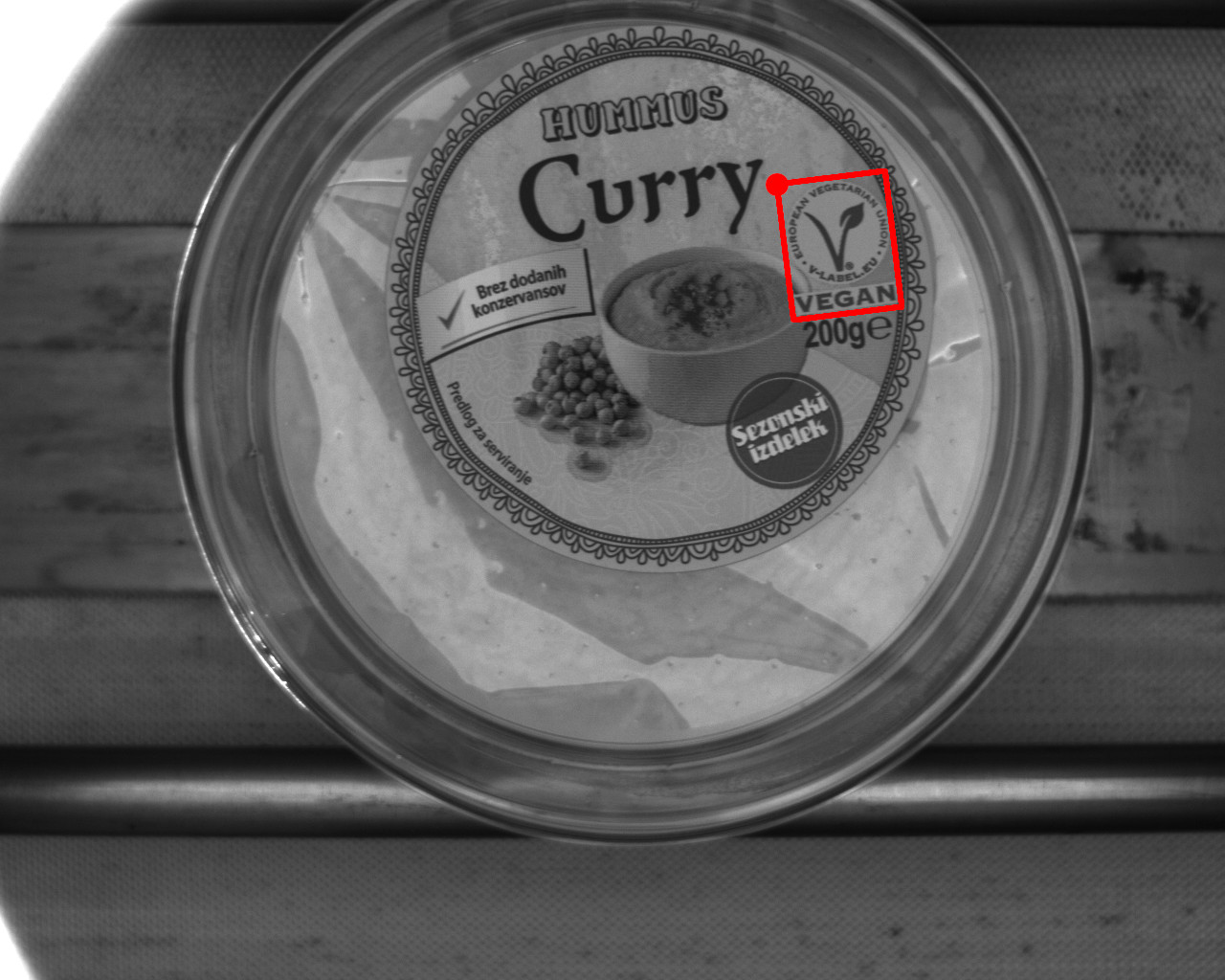}\includegraphics[width=.23\textwidth,cframe=gray!40]{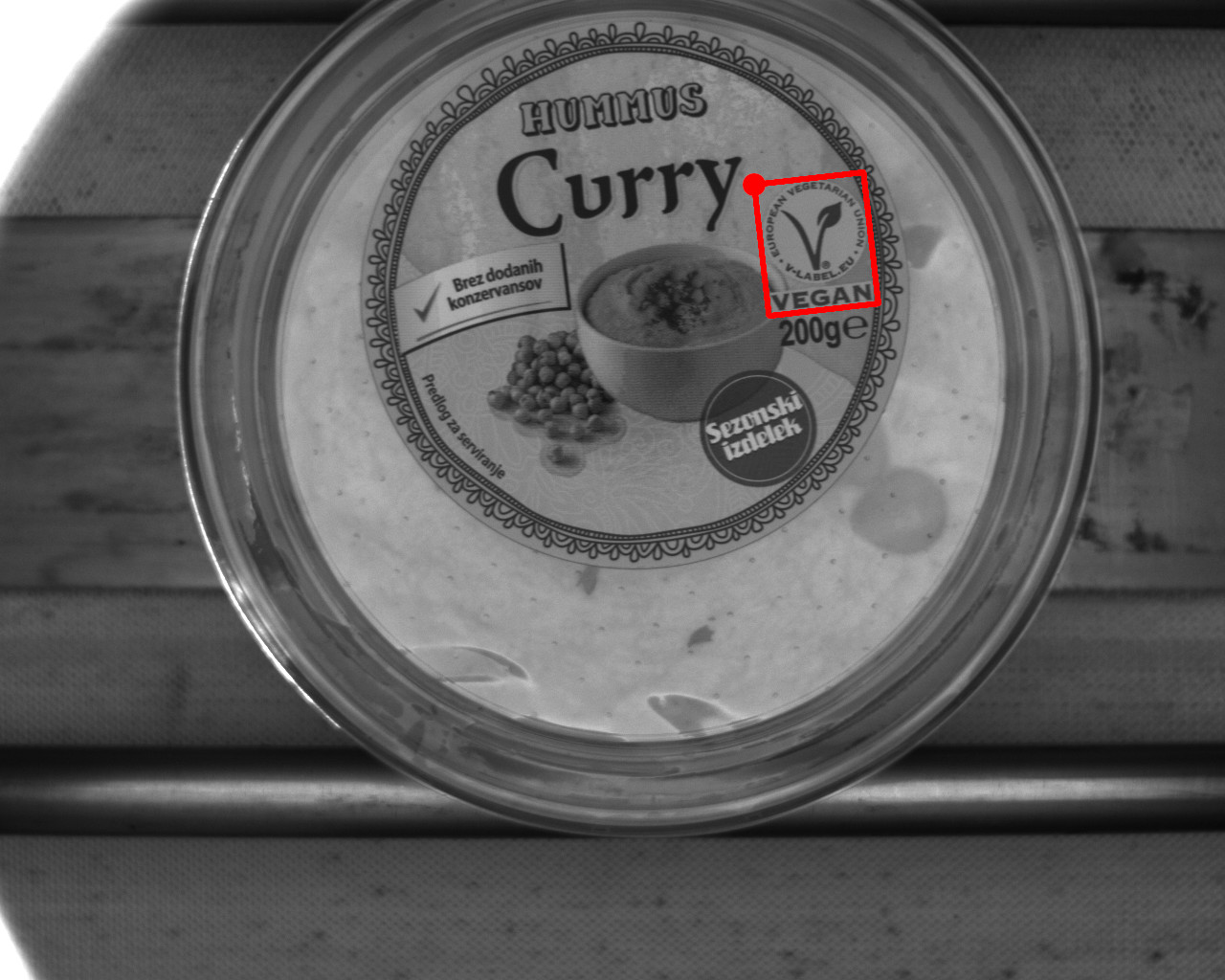}}\quad
    \subfloat[Tea, 31 images.]{\includegraphics[width=.23\textwidth]{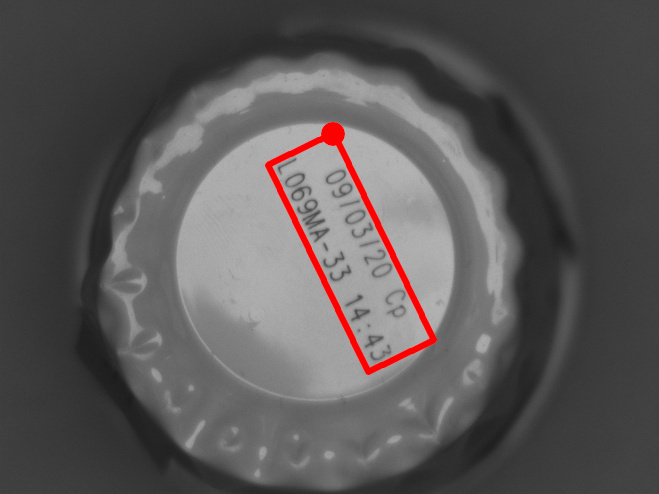}\includegraphics[width=.23\textwidth]{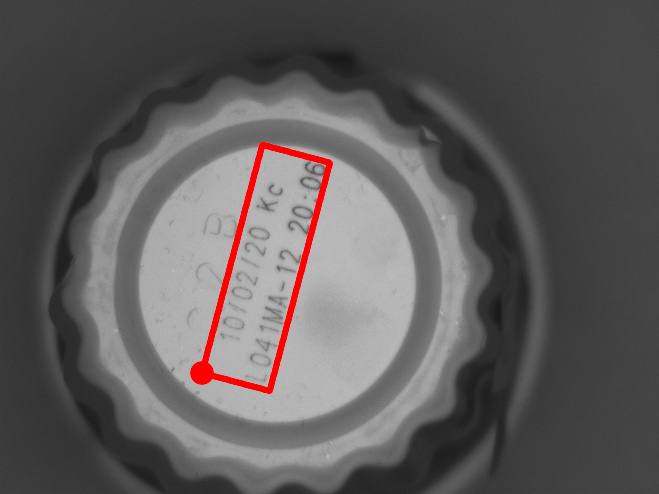}}\\
    \subfloat[Oil, 10 images.\label{fig:oil_ds}]{\includegraphics[width=.23\textwidth,cframe=gray!40]{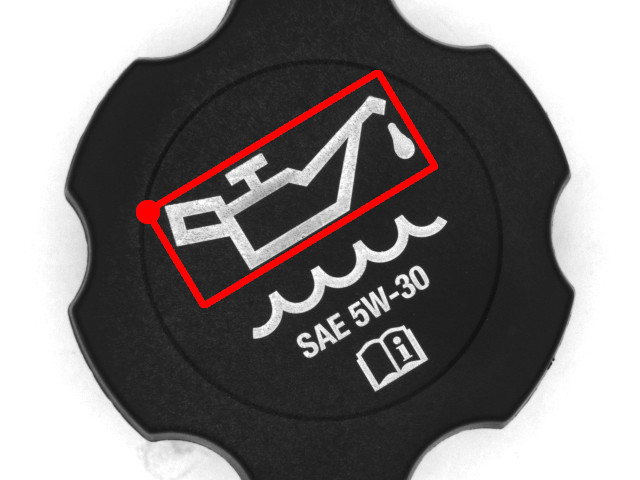}\includegraphics[width=.23\textwidth,cframe=gray!40]{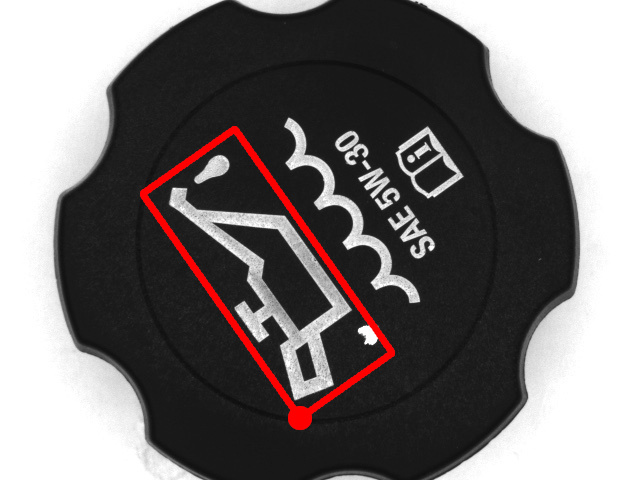}}\quad
    \subfloat[Pasta, 16 images.]{\includegraphics[width=.23\textwidth]{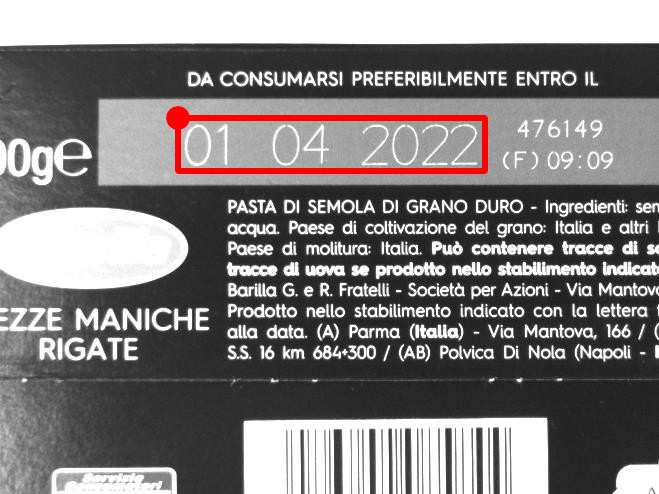}\includegraphics[width=.23\textwidth]{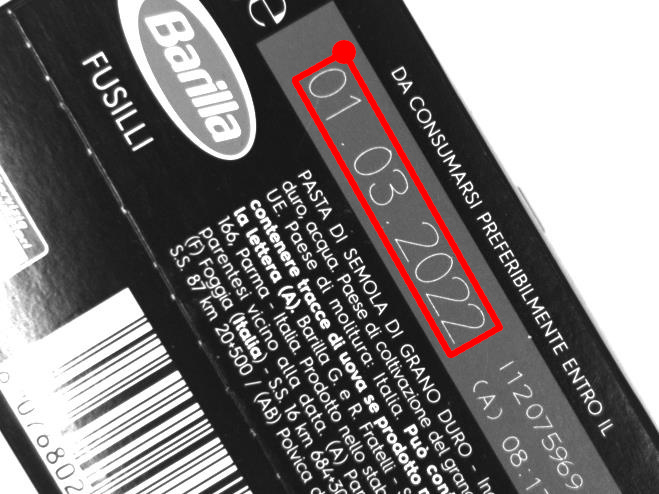}}\\
    \subfloat[Juice, 139 images.]{\includegraphics[width=.23\textwidth]{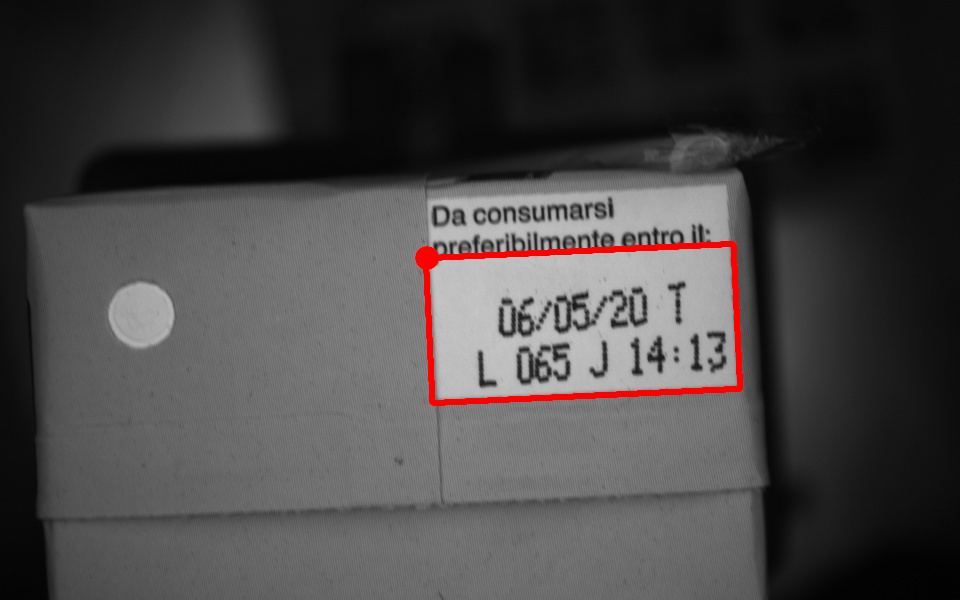}\includegraphics[width=.23\textwidth]{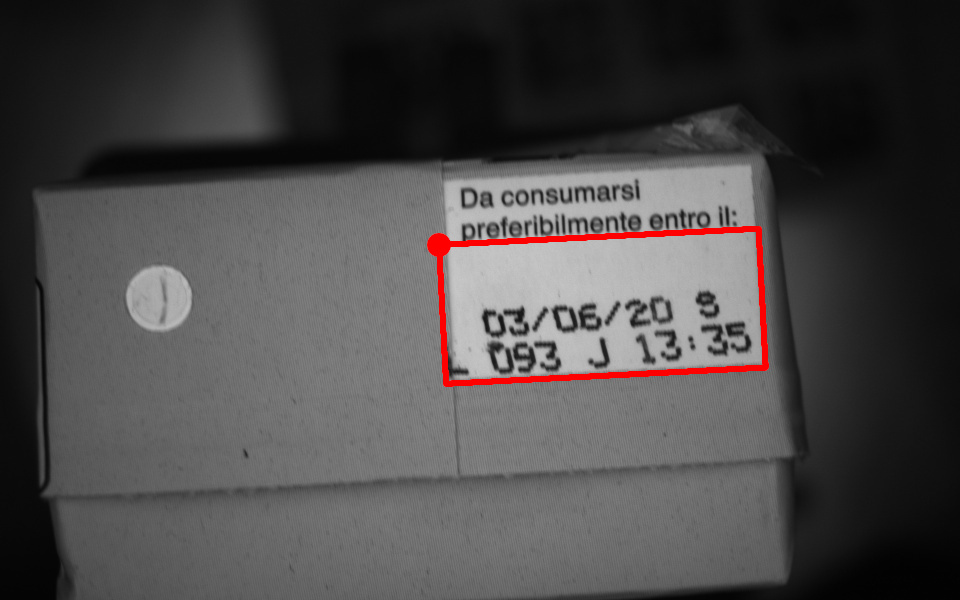}}\quad
    \subfloat[Ciocorì, 35 images. \label{fig:ciocory_ds}]{\includegraphics[width=.23\textwidth]{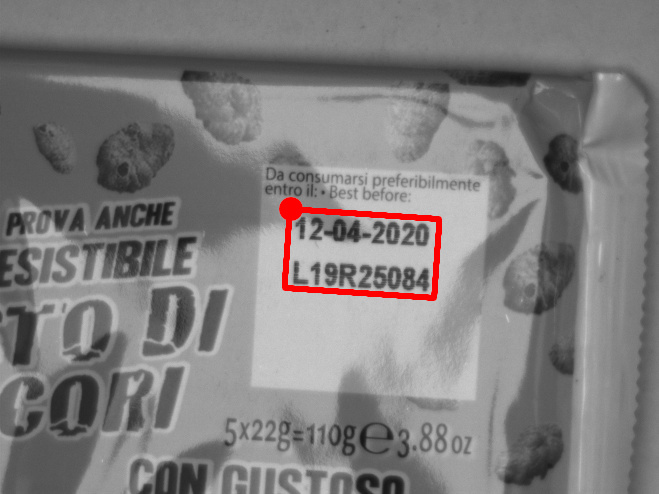}\includegraphics[width=.23\textwidth]{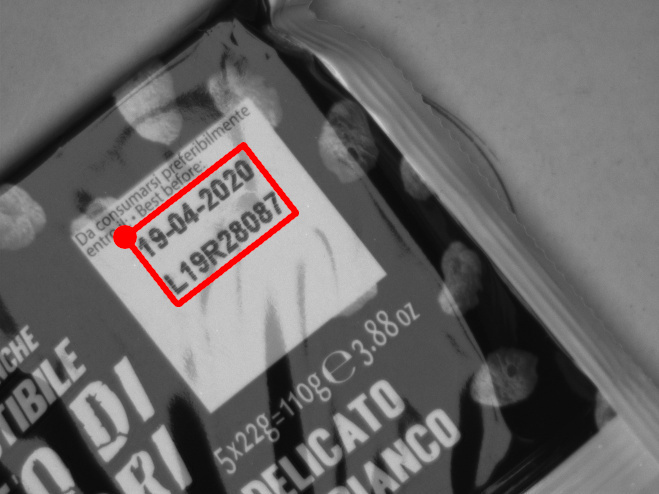}}\\
    \subfloat[LeadFrame, 11 images.\label{fig:leadframe_ds}]{\includegraphics[width=.23\textwidth]{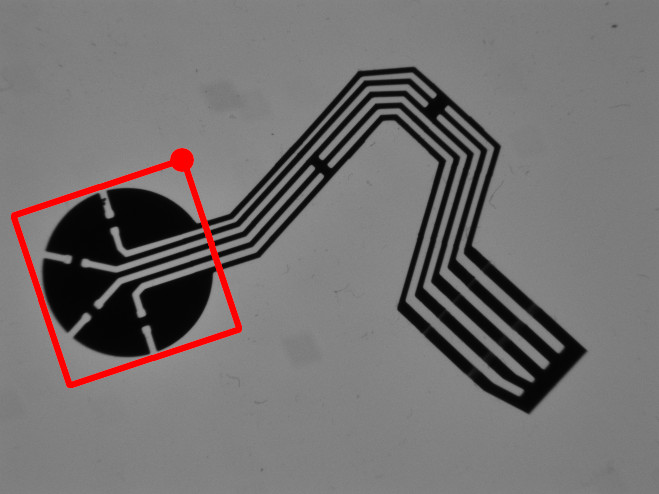}\includegraphics[width=.23\textwidth]{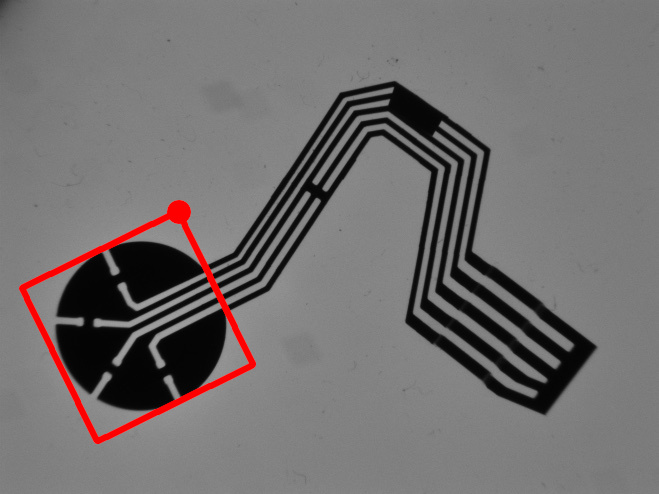}}\quad
    \subfloat[Caprice, 29 images.]{\includegraphics[width=.23\textwidth]{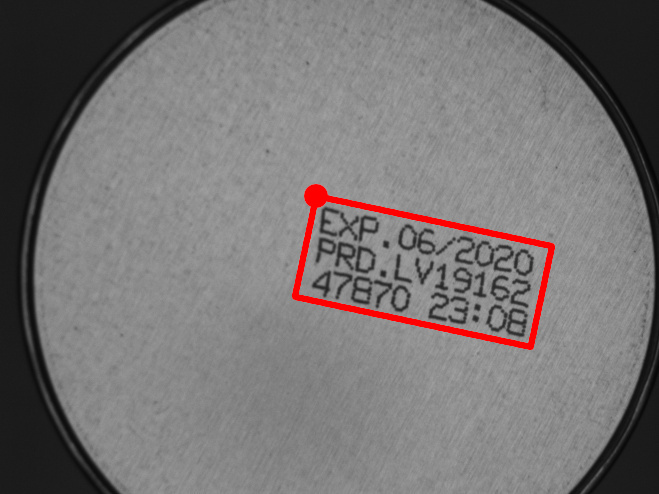}\includegraphics[width=.23\textwidth]{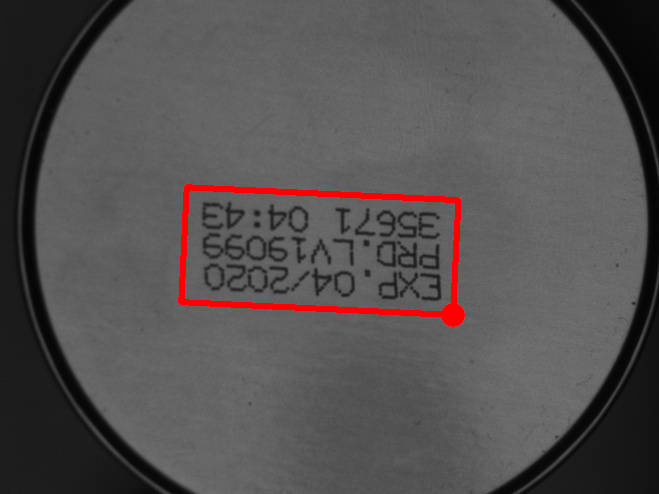}}\\
    \subfloat[Beer, 86 images.]{\includegraphics[width=.23\textwidth]{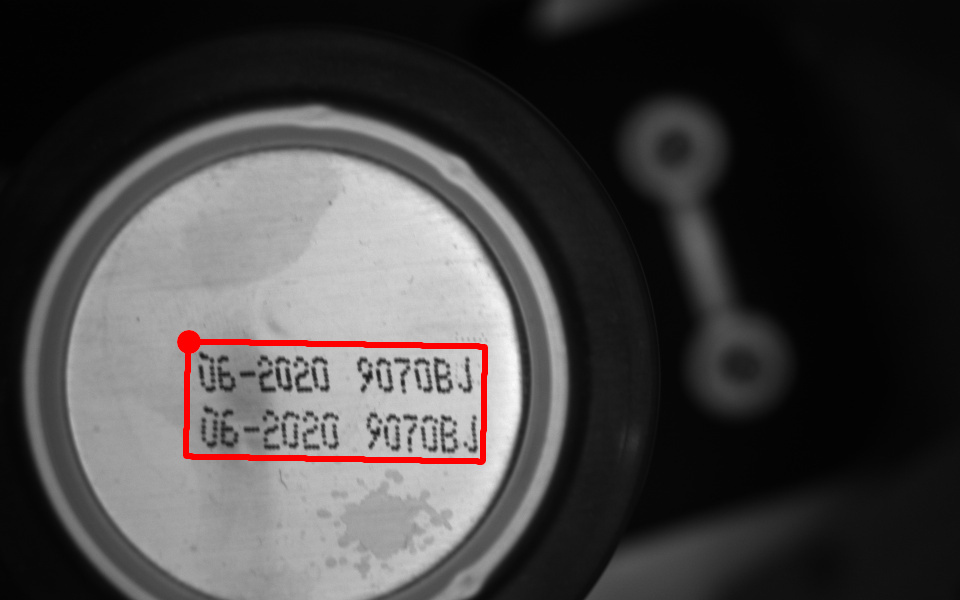}\includegraphics[width=.23\textwidth]{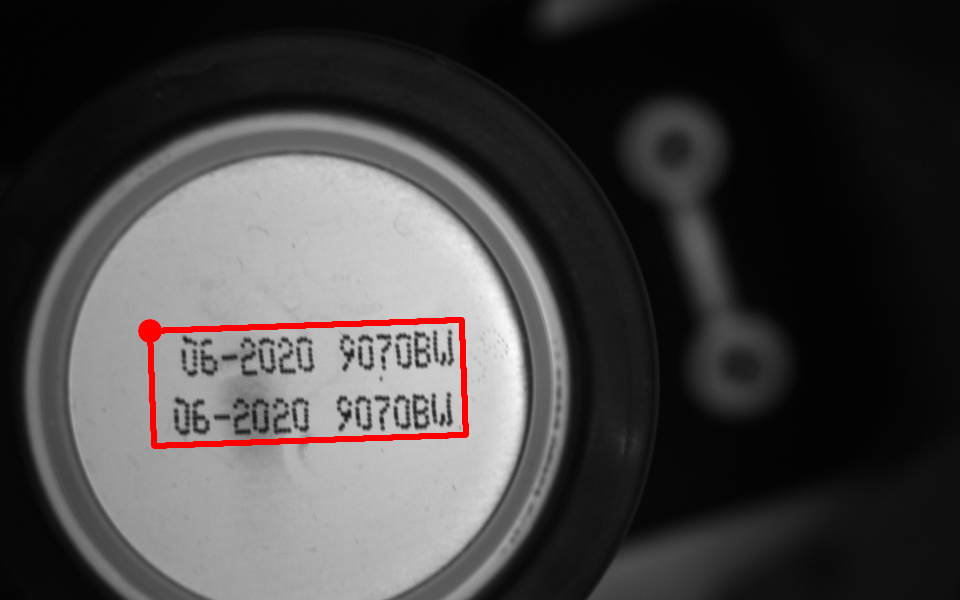}}
    \caption{Evaluation datasets: the ROI is depicted in red and the highlighted corner is the origin. Real datasets also include the total number of images.}
    \label{fig:eval_datasets}
\end{figure}

The datasets used to validate SAFFIRE are presented in \figref{fig:eval_datasets} with 2 samples from each alongside the total number of images available for each one. The datasets are not suited for a single task, but many can be applied: the main task is indeed reading the text to ensure it has been printed correctly, but also measurements (see \figref{fig:leadframe_ds}) and flaw detection (see \figref{fig:curry_ds} and \figref{fig:oil_ds}) can be carried out.

The images have been shuffled upon acquisition and the first three have been used for the training phase.
The results presented in \secref{sec:results} are computed on the whole datasets, including the images used in training.

\subsection{Evaluation Metrics}\label{sec:eval_metric}
The search capabilities of the trained models are evaluated by comparing the computed ROI with the ground truth one by means of $o\mathtt{IoU}$ between the two.
For each model the individual $o\mathtt{IoU}$'s are aggregated in a single plot showing the distribution of $o\mathtt{IoU}$ over the samples in the dataset by using a box plot.

The box in each box plot extends from the first $q_1$ to the third quartile $q_3$ of the distribution and, inside it, the median is highlighted.
Outside the box, the \say{whiskers} extend on the left to the lowest value in the distribution greater than $q_1 -1.5\cdot \mathrm{IQR}$, with the interquartile range (IQR) defined as $q_3-q_1$; similarly on the right to the highest value less than $q_3+1.5\cdot \mathrm{IQR}$.
Samples outside the range of the whiskers are considered outliers and plotted as single points.

\section{Results}
\label{sec:results}

\subsection{Noise Patterns and Filtering}\label{sec:results_noise}

As detailed in \secref{sec:path_search}, the path cost depends both on the number of overlaps of votes along the path and how well the ROI's are aligned, in the following we will show the importance of the second term.

\begin{figure}
    \centering
    \subfloat[Only overlaps.\label{fig:starcart_train_overlap_only}]{\includegraphics[width=.48\textwidth,cframe=gray!40]{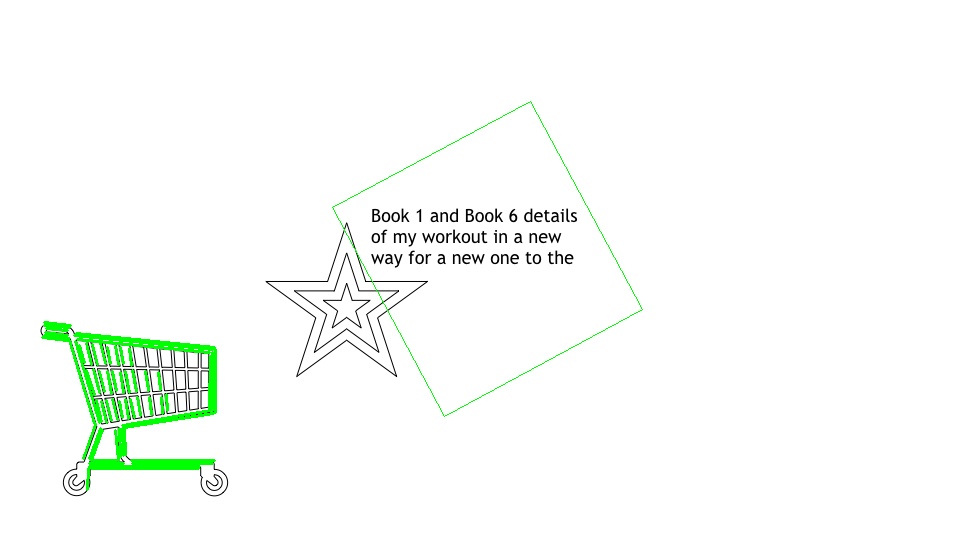}}\ %
    \subfloat[Overlaps and $o\mathtt{IoU}$.\label{fig:starcart_train_overlap_iou}]{\includegraphics[width=.48\textwidth,cframe=gray!40]{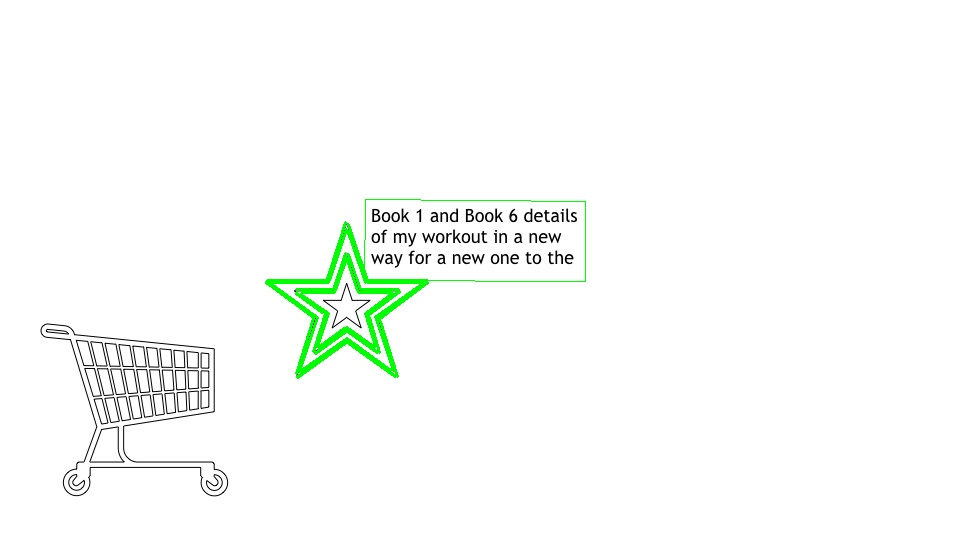}}
    \caption{Training over the StarCart dataset: the thick green lines are the segments of the model, the lighter ones delimit the ROI of said model.}
    \label{fig:starcart_train}
\end{figure}

Considering the StarCart dataset (see \figref{fig:starcart_ds}), it can be seen that both the cart and the star have a somewhat static position with respect to the ROI, however the orientation changes in the case of the cart.
We applied the training procedure using only segment based features due to the nature of the shapes, and report the results in \figref{fig:starcart_train}.
As depicted in \figref{fig:starcart_train_overlap_only}, considering only the overlaps results in taking the cart as the learned model due to the fact that its shape is formed by many more segments than the star is.
Adding the term based on ROI $o\mathtt{IoU}$ results instead in the desired model (see \figref{fig:starcart_train_overlap_iou}).

\subsection{Location Accuracy}

\begin{figure}
	\centering
    \includegraphics[width=\linewidth]{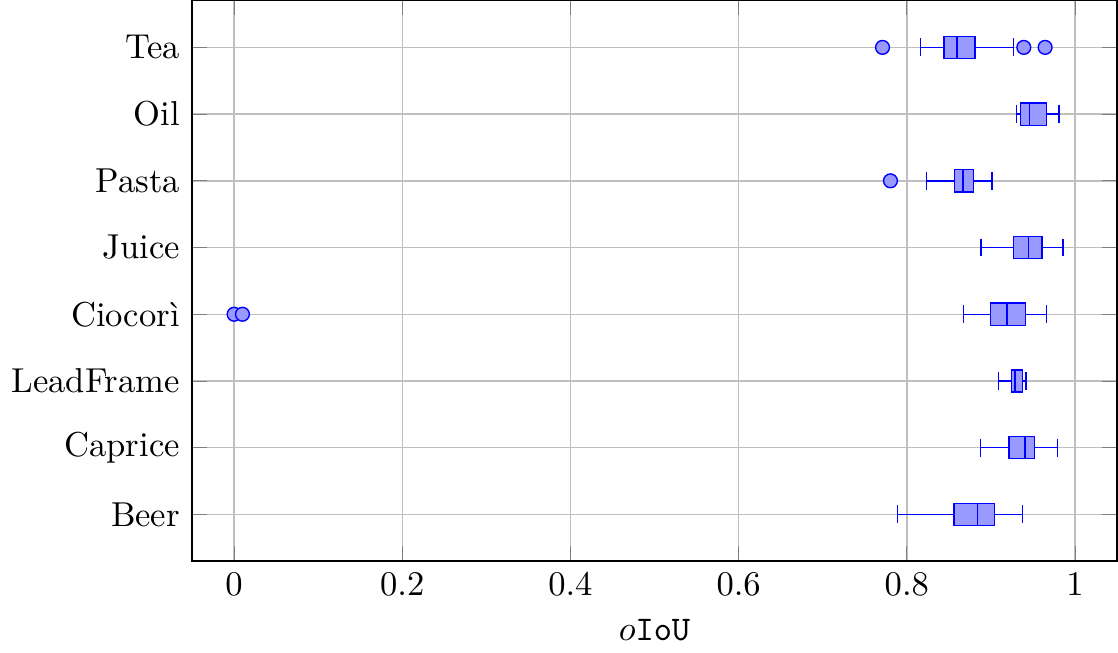}
	\caption{Distributions of $o\mathtt{IoU}$ of computed and ground truth ROI over several datasets.}
	\label{fig:iou_final_config}
\end{figure}

The search capabilities of SAFFIRE are evaluated over several different datasets (see \figref{fig:eval_datasets}) using $o\mathtt{IoU}$ as metric (see \secref{sec:eval_metric}) and reported in \figref{fig:iou_final_config}.

\subsection{Computation Time}

SAFFIRE has been implement to run on a Datalogic Memor 20,\footnote{\url{https://www.datalogic.com/eng/retail-manufacturing-transportation-logistics-healthcare/mobile-computers/memor-20-pd-869.html}} powered by a Snapdragon 600 CPU.
On this platform the train process takes just a few seconds and the search tens of milliseconds (with a few exceptions) as reported in \tbref{tb:times_final_config}. Search times are averaged over all images in the dataset and train times over 5 executions.

\begin{table}
	\centering
	\vspace{-5mm}
	\caption{Average times for training and search over several datasets.}
	\label{tb:times_final_config}
	\vspace{.4em}
	\begin{adjustbox}{center}
	\begin{tabular}{ll
		S[table-format=2.2,table-space-text-post = \,\si{\milli\second}]
		S[table-format=3.2,table-space-text-post = \,\si{\milli\second}]
		S[table-format=3.2,table-space-text-post = \,\si{\milli\second}]
		S[table-format=2.2,table-space-text-post = \,\si{\milli\second}]
		S[table-format=3.2,table-space-text-post = \,\si{\milli\second}]
		S[table-format=3.2,table-space-text-post = \,\si{\milli\second}]
		S[table-format=2.2,table-space-text-post = \,\si{\milli\second}]
		S[table-format=2.2,table-space-text-post = \,\si{\milli\second}]} \toprule
	              && \multicolumn{1}{c}{\textbf{Tea}} & \multicolumn{1}{c}{\textbf{Oil}} & \multicolumn{1}{c}{\textbf{Pasta}} & \multicolumn{1}{c}{\textbf{Juice}} & \multicolumn{1}{c}{\textbf{Ciocorì}} & \multicolumn{1}{c}{\textbf{LeadFrame}} & \multicolumn{1}{c}{\textbf{Caprice}} & \multicolumn{1}{c}{\textbf{Beer}} \\\midrule
	\multirow{2}{*}{\textbf{Train}} & \textbf{Time}
		&   1.30\,\si{\second}
		& 983.33\,\si{\milli\second}
		&   8.63\,\si{\second}
		&   3.13\,\si{\second}
		&   3.83\,\si{\second}
		& 896.10\,\si{\milli\second}
		&   1.24\,\si{\second}
		&   2.78\,\si{\second}\\
    & \textbf{Features} & \multicolumn{1}{c}{A} & \multicolumn{1}{c}{A} & \multicolumn{1}{c}{B} & \multicolumn{1}{c}{B} & \multicolumn{1}{c}{A} & \multicolumn{1}{c}{A} & \multicolumn{1}{c}{A} & \multicolumn{1}{c}{A} \\\midrule
	\multirow{2}{*}{\textbf{Search}} & \textbf{GHT}
		&   8.62\,\si{\milli\second}
		&   2.90\,\si{\milli\second}
		& 744.38\,\si{\micro\second}
		&   1.98\,\si{\milli\second}
		& 774.97\,\si{\micro\second}
		&   4.89\,\si{\milli\second}
		&   1.11\,\si{\milli\second}
		&  45.05\,\si{\milli\second} \\
	& \textbf{Total}
		&  37.05\,\si{\milli\second}
		&  36.20\,\si{\milli\second}
		& 106.21\,\si{\milli\second}
		&  99.16\,\si{\milli\second}
		&  35.28\,\si{\milli\second}
		&  38.89\,\si{\milli\second}
		&  32.32\,\si{\milli\second}
		&  86.70\,\si{\milli\second}
	\\\bottomrule
	\end{tabular}\end{adjustbox}
	\vspace{-3mm}
\end{table}

During development, SAFFIRE has been tested also on a personal computer powered by an Intel Core i5 7267U CPU.
The results over a different selection of datasets, which were mainly used during development, are reported in \tbref{tb:times_intelx86}.

\begin{table}
	\centering
	\vspace{-5mm}
	\caption{Average times for training and search over a few datasets on the development platform.}
	\label{tb:times_intelx86}
	\vspace{.4em}
	\begin{tabular}{ll
		S[table-format=3.2,table-space-text-post = \,\si{\milli\second}]%
		S[table-format=3.2,table-space-text-post = \,\si{\milli\second}]
		S[table-format=3.2,table-space-text-post = \,\si{\milli\second}]
		S[table-format=3.2,table-space-text-post = \,\si{\milli\second}]%
		S[table-format=2.2,table-space-text-post = \,\si{\milli\second}]
		S[table-format=2.2,table-space-text-post = \,\si{\milli\second}]
		S[table-format=2.2,table-space-text-post = \,\si{\milli\second}]%
		S[table-format=2.2,table-space-text-post = \,\si{\milli\second}]} \toprule
	         	&& \multicolumn{1}{c}{\textbf{Cans}}
	         	& \multicolumn{1}{c}{\textbf{Curry}}
	         	& \multicolumn{1}{c}{\textbf{Beer}} \\\midrule
	\multirow{2}{*}{\textbf{Train}} & \textbf{Time}
		&   4.92\,\si{\second}
		&   8.28\,\si{\second}
		&   3.99\,\si{\second}\\%
	& \textbf{Features} 
		& \multicolumn{1}{c}{B} 
		& \multicolumn{1}{c}{B} 
		& \multicolumn{1}{c}{A} \\\midrule
	\multirow{2}{*}{\textbf{Search}} & \textbf{GHT}
		&  19.40\,\si{\milli\second}
		&  41.59\,\si{\milli\second}
		&  26.53\,\si{\milli\second}\\
	& \textbf{Total}
		& 126.76\,\si{\milli\second}
		& 216.23\,\si{\milli\second}
		&  35.95\,\si{\milli\second} \\\bottomrule
	\end{tabular}
	\vspace{-3mm}
\end{table}

\section{Discussions}
\label{sec:discussion}

The results in \figref{fig:iou_final_config} shows that SAFFIRE can obtain an high accuracy in identifying the ROI.
From our experiments and by means of a qualitative evaluation of results, we found that even with $o\mathtt{IoU}$ of about 0.85, localization of the ROI can be considered accurate.
This is caused by using as model a ROI slightly larger than those used for training and as ground truth, due to how it is constructed.
We also note some failures for Ciocorì denoted by the points around 0, likely caused by the reflective surface of the packaging (see \figref{fig:ciocory_ds}).

\tbref{tb:times_final_config} indeed shows that the search time is fast, but by isolating the identification of the ROI, by means of the GHT and the ROI content descriptor to choose the right candidate, it emerges how our approach is extremely fast and most of the time is taken by feature extraction and descriptor matching.
Indeed, the fastest features are preferred as SAFFIRE prefers choosing the ones with the least computational time even if this means sacrificing a bit of accuracy (see \secref{sec:best_feature_type}).
The results in \tbref{tb:times_intelx86} are provided as a mean of comparison of running SAFFIRE on an non-embedded device and, as such, they lack some optimizations that led to the final results.

We also note that the SAFFIRE framework is extremely general and independent on the specific technical choices presented in this work:
\begin{itemize}
    \item Any feature descriptor can be used;
    \item Other techniques can be used instead of RANSAC to compute the nodes of the training graph, such as the GHT;
    \item Arbitrary spatial transforms can be employed depending on the geometry of the problem;
    \item Any graph-search algorithm can be used to find the optimal model;
    \item Many image descriptors can be used to encode the ROI image content.
    \item The rectangular ROI in this study is not a limitation and can be replaced with arbitrary shapes with a defined notion of \say{direction}, in order to simplify the analysis task following the ROI identification, e.g., to force text to be upright, with this definition even circles can be used.
\end{itemize}

The algorithm can also be adapted to work in a \say{fully unsupervised} fashion not requiring the provision of a ROI. This is simply achieved by adjusting the path cost in the graph to depend only on the number of overlaps, thus discarding the alignment of the ROI's. This way, SAFFIRE will look for the most stable pattern in the image sequence with the highest number of features. For instance, this could be sufficient to align images of similar items acquired in different positions, but care must be taken as it can results in cases like that presented in \secref{sec:results_noise}, in which relying only on feature overlaps results in a bad alignment of ROI's.

\section{Conclusions}
\label{sec:conclusions}
We have introduced a new algorithm, called SAFFIRE, for unsupervised identification of a stable anchor pattern from a set of images. Such an anchor is used to remove pose variations between samples before further processing. Data normalization is a fundamental pre-processing step for many computer vision and machine learning tasks.

The use of SAFFIRE was shown in the context of machine vision systems for industrial automation, where it represents a disruptive element enabling the implementation of an automatic ROI finding mechanism in an unsupervised fashion. 

\section*{Intellectual Property}
Patent application pending.
%
%
%
\bibliographystyle{splncs04}
\bibliography{saffire}





\end{document}